\documentclass[runningheads]{llncs}

\usepackage{eccv}

\usepackage{eccvabbrv}
\usepackage{graphicx}
\usepackage{booktabs}
\usepackage{amsmath}
\usepackage[section]{placeins}
\usepackage[accsupp]{axessibility}
\usepackage[pagebackref,breaklinks,colorlinks,citecolor=eccvblue]{hyperref}

\newcommand{\citep}[1]{\cite{#1}}
\newcommand{\citet}[1]{\cite{#1}}

\begin{document}

\title{Oh Deer, How Should I Handle This? Seasonal Priors for Selective Wildlife Annotation and Classification}
\titlerunning{Oh Deer, How Should I Handle This?}

\newcommand{\equalcontrib}{\textsuperscript{*}}

\author{
  Hugo Markoff\inst{1}\equalcontrib \and
  Christoph Praschl\inst{2}\equalcontrib \and
  Anton Hjalte J{\o}rgensen\inst{1} \and
  Christian Emil Mogensen\inst{1} \and
  Mathias Bech Skadhauge\inst{1} \and
  Sara Beery\inst{3} \and
  Michael {\O}rsted\inst{1} \and
  David C. Schedl\inst{2}
}

\authorrunning{H. Markoff et al.}

\institute{
  Aalborg University, Department of Chemistry and Bioscience, Aalborg, Denmark
  \and
  University of Applied Sciences Upper Austria, Hagenberg, Austria
  \and
  Massachusetts Institute of Technology, Cambridge, MA, USA
}

\maketitle

\begingroup
\renewcommand{\thefootnote}{*}
\footnotetext{Corresponding authors (equal contribution): Hugo Markoff (khbm@bio.aau.dk), Christoph Praschl (christoph.praschl@fh-hagenberg.at)}
\endgroup

\begin{abstract}
Fine-grained wildlife classification in aerial imagery is limited not only by model performance, but also by unreliable labels: animals occupy few pixels, key visual cues vary seasonally, and modality-specific evidence can be ambiguous. We study adult-male identification in red deer \textit{(Cervus elaphus)}, where the antler cycle defines predictable windows of reliable evidence for both annotation and prediction. Using 7,295 RGB-only, thermal-only, and matched RGB+thermal crop sets from low-altitude UAV surveys, labeled by three annotators, we show that seasonal structure links (I) annotation quality, (II) downstream classification, and (III) selective prediction. Matched RGB+thermal review resolves more samples than either single modality, recovering majority-male labels otherwise missed by RGB or thermal alone, in human-based as well as model-based classification. Months with high annotator abstention also show lower classifier confidence, and soft seasonal priors mainly benefit the season-limited thermal view. Uncertainty-band abstention further raises covered accuracy to 98.9\%, though at reduced coverage and with deferral that falls disproportionately on males. Overall, a biologically grounded seasonal calendar predicts where annotation and prediction are unreliable, and can guide both annotation protocol design and modality weighting.
\end{abstract}

\vspace{-0.3em}
\noindent

\section{Introduction}
\label{sec:intro}

Accurate estimates of wildlife abundance and demographic composition are central to ecological monitoring and population management. In particular, information about species, sex, and age supports analyses of recruitment, survival, reproduction, and long-term population dynamics~\citep{clutton1982reddeer}. Low-altitude uncrewed aerial vehicle (UAV) surveys are increasingly attractive because they can cover large areas efficiently while reducing disturbance compared to ground-based observations~\citep{hodgson2018drones}. However, aerial imagery also introduces a major limitation: animals often occupy only a small image region, making fine-grained interpretation substantially harder than coarse tasks such as detection or counting~\citep{kellenberger2018detecting,burke2019thermal}. As a result, most aerial wildlife studies still focus primarily on localization, detection, and abundance estimation, while fine-grained classification remains much less established~\citep{chretien2016,bondi2020birdsai,weinstein2022general,zabel2023}.

A central challenge is that fine-grained classification depends on labels that are themselves hard to assign reliably. Subtle traits may be weak, partially occluded, seasonally absent, or inconsistently interpreted, and not all observations support reliable detailed identification~\citep{kays2022which,mcKibben2021linking}. In aerial imagery, especially thermal, this is amplified by low effective resolution, sensor motion, and changing environmental conditions. Fine-grained classification can therefore fail already at dataset construction, before a model can learn the distinctions, because humans must label them first, which is especially problematic under class imbalance where modest rates of systematic labeling error distort downstream decision boundaries~\citep{northcutt2021labelerrors}.

Among fine-grained attributes, sex most directly constrains demographic inference: adult-male counts and sex ratios underpin harvest quotas, rut-season management, and recruitment estimates, and unlike species identity they cannot be narrowed by a habitat or range prior. In sexually dimorphic species, the visual evidence for sex can itself be strongly seasonal, making them an ideal setting in which to ask whether the evidence suffices at all. We therefore treat aerial sex classification first as a problem of label quality and \emph{evidence sufficiency} rather than of sample acquisition: the question is whether a crop shows enough to support a sex label at all. Taking species-level annotation as given, we focus on adult-male identification in red deer, where the primary cue, antler visibility, changes predictably across the annual cycle and differs between RGB and single-channel thermal-intensity imagery. We encode this cycle as a \emph{soft prior} over how informative each modality is in a given month. It is soft because it never assigns or overrides a label, and never shifts a prediction at inference; it only states how much to trust each modality. Only in a single ablation do we test it in a non-soft setup for weighting training samples (\autoref{subsec:priors}).

Our imagery comes from low-altitude UAV flights with a co-mounted RGB camera and non-radiometric thermal sensor at a near-nadir viewing angle. Using aligned crop sets labeled under RGB-only, thermal-only, and matched conditions, we ask how much resolved supervision matched review recovers over single-view annotation (\autoref{subsec:annotation_results}); how much each feature view contributes once supervision is fixed to a common matched label pool, and when abstention beats forcing a binary prediction on weak evidence (\autoref{subsec:modality_results}--\ref{subsec:classifier_families}); and when soft seasonal priors explain or improve modality-specific behavior (\autoref{subsec:priors}). These share one thread: the antler cycle creates a seasonal calendar of evidence that predicts both when annotators will struggle and when model confidence will be low.

\section{Related Work}
\label{sec:related}

This paper sits at the intersection of four topics: aerial wildlife monitoring, label quality under ambiguous evidence, representation learning for fine-grained classification, and seasonally grounded ecological priors.

\subsection{Aerial Wildlife Monitoring and Fine-Grained Interpretation}

While most aerial wildlife studies focus on detection and counting (see \autoref{sec:intro}), a few show that sex and age can be inferred under favorable conditions, for example through antler visibility in RGB or manual interpretation of thermal drone data~\citep{ito2022antler,larsen2023thermal}. These studies share a limitation: reliable fine-grained labels depend on whether the relevant traits are actually visible. What remains unexplored is whether a known biological seasonality calendar can anticipate which modality will be informative at which time of year, and be built into the annotation protocol from the start.

\subsection{Label Quality, Selective Annotation, and Ecological Priors}

Interactive review tools and active learning reduce annotation burden in image-based survey workflows~\citep{kellenberger2019halfpercent,kellenberger2020aide,settles2009activelearning}. Our focus is narrower: not how to request more labels, but how to avoid turning seasonally ambiguous observations into hard label noise. This connects to label-quality work on systematic annotation error in imbalanced datasets~\citep{northcutt2021labelerrors}, and to selective prediction, where abstention is preferable when confidence is low~\citep{geifman2017selective}. Our seasonal calendar is complementary to, rather than a replacement for, post-hoc confidence-based filtering: it anticipates ambiguity before annotation begins, whereas the abstention rule we evaluate at inference time remains purely confidence-based and month-agnostic.

\subsection{Hierarchical and Representation-Based Fine-Grained Classification}

Fine-grained wildlife labels naturally form a hierarchy, in which demographic traits such as sex and age are inferred only once visibility and species identity are sufficiently supported; this improves consistency in ecological label spaces and supports robust coarse-level predictions when fine-grained decisions are uncertain~\citep{elhamod2022hierarchy,bjerge2023hierarchical,weinbach2025fjordvision}. We take species-level annotations as given and address the sex-labeling step below them. Pretrained Vision Transformers and self-supervised models such as DINO provide strong embeddings for biological imagery~\citep{dosovitskiy2020vit,oquab2023dinov2,simeoni2025dinov3} and can preserve ecologically meaningful intra-species variation~\citep{markoff2026vit}, while metric learning adapts those embeddings to subtle class differences~\citep{schroff2015facenet,musgrave2020metric,miele2021revisiting}. In aerial settings, RGB and thermal representations are especially relevant because the modalities offer complementary cues for both annotation and classification~\citep{krishnan2023fusion}.

\subsection{Seasonal Antler Priors for Aerial Male Identification}

Red deer are a suitable test case for ecologically grounded fine-grained classification because the primary sex cue, antler presence, varies predictably across the year. In European populations, antlers are cast in late winter to early spring, regrow under velvet over roughly four to five months, and the velvet is then shed to reveal hard, clean antlers by late summer before the autumn rut~\citep{lincoln1992antlercycle,bils2023,waldwissen_reddeer}. Thermography uses a thermal-infrared sensor to record an animal’s surface temperature by detecting emitted infrared radiation rather than reflected visible light. This imaging methodology shows that velvet-antler temperatures in red deer peak during early and mid growth and decline toward late growth as ossification progresses~\citep{bowers2010}. The same measurements on other cervids (the deer family, including elk and roe deer) further suggest a marked temperature drop near velvet shedding, consistent with reduced blood supply as the velvet phase ends~\citep{potrapeluk2021}. Drone studies additionally show that antlers may remain recognizable in thermal imagery after velvet loss, even outside the main high-temperature growth phase~\citep{larsen2023thermal,ito2022antler}.

\begin{figure}[htbp]
    \centering
    \includegraphics[width=0.5\linewidth]{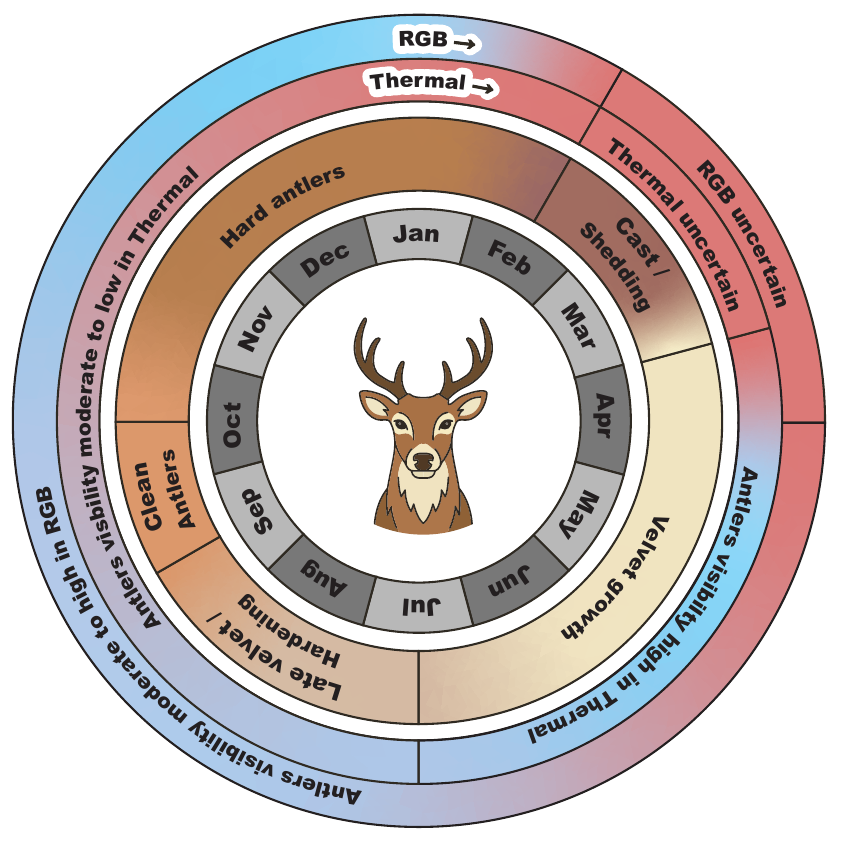}
    \caption{Seasonal visibility of red deer antlers in aerial RGB and thermal imagery. The circular diagram illustrates the annual cycle (inner ring, months) and corresponding antler states (middle ring). The split outer ring indicates the uncertainty in detecting antlers in the two imaging modalities.}
    \label{fig:antler_cycle}
\end{figure}

These observations motivate soft prior expectations for antler-based sexing (cf. \autoref{fig:antler_cycle}). The least reliable period is the casting window around late February to March, sometimes extending into early April, when antlers may be absent or only begin to regrow~\citep{bils2023,bds_antlercycles}. Thermal imagery should be most informative during early-to-mid velvet growth, when vascular activity is high and heat-based contrast is strongest~\citep{bowers2010}. RGB cues should become increasingly informative after hardening and cleaning, when antlers are visually prominent. During the transition from late velvet to early cleaning (approximately late July to August), the two modalities are likely complementary rather than redundant~\citep{potrapeluk2021,larsen2023thermal,waldwissen_reddeer}. We use these per-modality prior windows primarily as an annotation-quality diagnostic, and secondarily test them as a training signal, connecting the biological literature to concrete protocol decisions.
\section{Method}
\label{sec:methods}

\subsection{Dataset, Task, and Evaluation Design}

\begin{figure}[htbp]
\centering
\includegraphics[width=0.8\linewidth]{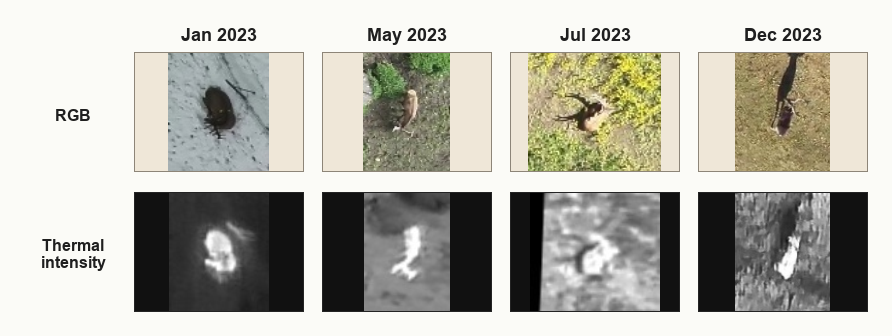}
\caption{Illustrative nadir RGB and thermal crop pairs from matched-view male examples in January, May, July, and December.}
\label{fig:season_crop_examples}
\end{figure}

We use red deer examples from a public aerial dataset (\url{https://doi.org/10.5281/zenodo.19034999}) containing synchronized RGB and non-radiometric thermal imagery collected in Austria across multiple years and months. The animals are observed from an approximately nadir UAV perspective, so sex cues must be inferred from small overhead crops rather than side views. Representative crop pairs are shown in \autoref{fig:season_crop_examples}. We analyze 7,295 aligned crop sets across the nine calendar months with sufficient coverage (January, February, March, April, May, July, September, November, and December); the remaining months lack enough flights to support a monthly breakdown. Surveys were flown at $30$--$60$\,m above ground level, depending on local vegetation height, with DJI M30T and M3T payloads. Paired RGB and thermal crops both have median size $72 \times 67$ pixels (interquartile range roughly $64$--$82$ wide, $58$--$77$ high), so an adult red deer spans only a few tens of pixels in either modality. This limited spatial support helps explain why antlers may be weakly visible and why subtle sex cues are seasonally hard even for human annotators. Throughout, \emph{matched} describes the \emph{data} -- one RGB and one thermal crop of the same animal, registered to each other as closely as pixel-level alignment of the two sensors allows -- while \emph{fused} describes what a model then does with such a pair.

Since antlers are the main visual cue for distinguishing males from females, and juveniles lack antlers as well, this feature cannot be used to differentiate females from juveniles. Because of this, our analyses target adult-male versus \sloppy{female/juvenile} classification. We keep \textit{uncertain} and \textit{occluded} as explicit abstention states instead of forcing them into a binary sex class. For cross-modal comparison, we define a common evaluation pool using the majority-vote label across all annotators from matched RGB+thermal review whenever that review resolves the sample. This keeps RGB-only, thermal-only, and matched feature views on identical target labels. This pool is a reference standard rather than externally validated ground truth: the same three annotators produce it, only with more evidence available, so it may inherit biases shared across views. We therefore report differences between views as label changes, not as corrections. Evaluation sets are formed at the flight level so that aligned held-out flights are shared across models and correlated crops from the same flight do not leak across training and evaluation.

\subsection{Three-View Annotation Protocol}

Each sample is inspected in three ways: RGB-only, thermal-only, and a matched RGB+thermal view. The thermal-only view uses the native single-channel intensity crop rather than a false-color palette. The three annotators all have academic ecology backgrounds. Each independently assigns one of four labels in each view: male, female/juvenile, uncertain, or occluded. The interface showed the capture month but not the exact flight date, so annotator judgments are not blind to season. Majority vote defines the label for that view: a view is \emph{resolved} if at least two annotators agree on male or female/juvenile, \emph{uncertain} or \emph{occluded} if at least two agree on that state, and \emph{no majority} if all three annotators disagree. No-majority samples are excluded from the evaluation pool rather than being broken by a tie rule, and they are reported explicitly in \autoref{tab:annotation_summary} so that the four outcome columns and the pool size reconcile. The matched majority vote supplies the shared supervision target for the common evaluation pool benchmarks, while the single-view votes are retained to analyze modality-specific resolution and disagreement.

\subsection{Seasonal Prior Bins}

Using the flight date and the red deer antler cycle, each sample is assigned a soft prior bin for each modality: RGB is expected to be most informative during hard-antler periods, thermal during velvet growth, while casting and early regrowth are treated as uncertain. For matched analysis the two single-modality priors are combined. A bin states expected \emph{evidence quality}, not an expected class, and we put it to exactly two uses: flagging the seasons in which single-modality review is unreliable (\autoref{subsec:annotation_results}), and setting training weights that favor reliable examples (\autoref{subsec:models}). Everywhere else it is read-only; in particular, the abstention rule never sees the month.

\subsection{Representations and Compared Models}
\label{subsec:models}

Each crop is encoded with a frozen DINOv3 ViT-H+ backbone~\citep{simeoni2025dinov3}, yielding a 1280-dimensional embedding per modality:
\[
z_i^{\text{rgb}} = E(x_i^{\text{rgb}}), \qquad
z_i^{\text{th}} = E(x_i^{\text{th}}).
\]

Because the frozen DINO backbone expects three input channels, the thermal crop is converted to pseudo-RGB by copying the same intensity image into all three channels before encoding.

For paired inputs, our primary fused representation concatenates the synchronized modality embeddings:

\[
z_i^{\text{mm}} = [z_i^{\text{rgb}}; z_i^{\text{th}}].
\]

Both views pass through the same RGB-pretrained backbone, so the RGB-versus-thermal ordering holds for this frozen representation rather than for the modalities in general (\autoref{subsec:classifier_families}). As an early-fusion control we also evaluate channel-replacement inputs, which substitute the thermal intensity image for one of the three RGB channels before DINO encoding,
\[
x_i^{R\leftarrow \text{th}} = [T_i, G_i, B_i], \qquad
x_i^{G\leftarrow \text{th}} = [R_i, T_i, B_i], \qquad
x_i^{B\leftarrow \text{th}} = [R_i, G_i, T_i].
\]

Our reference classifier is a linear SVM trained on the resolved subset,
\[
m(x) = w^\top z(x) + b,
\]
where $m(x)$ is the signed decision margin. With sample-specific weights $\alpha_i$, the training objective is
\[
\min_{w,b} \ \frac{1}{2}\lVert w \rVert_2^2 + C \sum_{i \in \mathcal{R}} \alpha_i \, \max\!\bigl(0, 1 - y_i (w^\top z_i + b)\bigr),
\]
where $\mathcal{R}$ is the resolved training set and $y_i \in \{-1,+1\}$ denotes female/juvenile versus male. This linear head is the main benchmark because it keeps the comparison focused on representation quality rather than classifier capacity.
In the unweighted baseline, $\alpha_i=1$ is used for every training sample; in the seasonal-weighting follow-up described in \autoref{subsec:priors}, $\alpha_i$ is adapted based on the antler cycle.

When comparing feature views, the resolved matched majority-vote pool is fixed across RGB-only, thermal-only, late-fusion, and channel-replacement models. This prevents label-availability differences from being mistaken for representation differences. For the seasonal-weighting ablation, each training weight $\alpha_i$ is set by the sample's soft-prior bin: weights above 1 up-weight samples from periods where the relevant cues should be clearly visible; weights below 1 down-weight samples from uncertain periods where cues may be absent or weak.

To test whether the cross-modal conclusion depends on the downstream model family, we additionally evaluate two alternatives. The first is triplet metric learning on the frozen embeddings,
\[
\mathcal{L}_{\text{triplet}} = \max\!\bigl(0, \lVert f_\theta(a) - f_\theta(p) \rVert_2 - \lVert f_\theta(a) - f_\theta(n) \rVert_2 + \tau\bigr),
\]
with margin $\tau = 0.5$. At inference time, a sample is assigned to the class with the nearest class centroid in the learned space,
\[
\hat{y}(x) = \arg\min_{c \in \{-1,+1\}} \lVert f_\theta(z(x)) - \mu_c \rVert_2^2,
\]
where $\mu_c$ is the training-set centroid of class $c$. The second is an image-space Ultralytics YOLO classifier trained directly on RGB or thermal crops rather than on frozen embeddings. Protocols differ: the SVM is reported over grouped five-fold held-out-flight splits, the triplet and YOLO checks over a single grouped seed-42 split. Both split definitions are released with the dataset. Since SVM and triplet share the same frozen features, their agreement tests the classifier head only; YOLO is the sole comparison that varies the encoder.

\subsection{Selective Prediction}

For each model family, we form a scalar male score $s(x) \in [0,1]$. For the linear SVM, $s(x) = \sigma\bigl(m(x)\bigr)$ is the sigmoid of the raw signed margin, with no Platt scaling or other post-hoc calibration. For the triplet model, $s(x)$ is derived from the distances to the two class centroids, and for YOLO it is the network's own male class probability; in neither of these two cases is a sigmoid applied. Selective prediction uses a symmetric uncertainty band over $s(x)$: for a half-width $\delta \ge 0$, samples with $|s(x)-0.5| \le \delta$ are abstained rather than forced into a binary class~\citep{geifman2017selective}, giving the selective classifier
\[
\hat{y}_\delta(x)=
\begin{cases}
\text{female/juvenile}, & s(x) < 0.5-\delta,\\
\text{abstain}, & 0.5-\delta \le s(x) \le 0.5+\delta,\\
\text{male}, & s(x) > 0.5+\delta.
\end{cases}
\]

These scores are not calibrated to a common scale, so the same band means different things across families: for the SVM, $0.35 \le s(x) \le 0.65$ is a raw margin of only $|m(x)| \le 0.619$. The band is therefore one operating point, and views should be compared at equal coverage from \autoref{fig:common_pool_modalities}. We report coverage and covered accuracy, which separate confident classification on supported evidence from explicit deferral on ambiguous cases.

\section{Results}
\label{sec:results}

\subsection{Human modality annotation, review and evaluation data pool}
\label{subsec:annotation_results}
Each crop is independently reviewed under single and matched conditions by three annotators, and the supervision recovered by each view is quantified before any classifier is trained. \autoref{tab:annotation_summary} shows that matched annotation resolves 52.8\% of the crop sets, compared with 32.0\% for RGB-only and 27.3\% for thermal-only review. The gain is not only more labels, but more consistent ones: matched review returns a majority-male label for 99 crops that RGB-only annotation left unresolved or called female/juvenile, and for 249 that thermal-only annotation did, while moving 22 RGB-only and 18 thermal-only male calls away from male once the paired view is available. Because the matched vote comes from the same three annotators, these are label changes rather than verified corrections.

\begin{table}[htbp]
\centering
\caption{Majority-vote annotation outcomes on 7,295 aligned crop sets. Matched review resolves 52.8\% of crop sets against 32.0\% for RGB-only and 27.3\% for thermal-only, so showing annotators both sensors recovers far more usable supervision than either view alone. `Resolved' denotes the share labeled as male or female/juvenile rather than uncertain or occluded. `No majority' counts crops on which all three annotators disagreed, so that each row sums to 7,295.}
\label{tab:annotation_summary}
\scriptsize
\setlength{\tabcolsep}{4pt}
\begin{tabular}{@{}l c r r r r r@{}}
\toprule
\multicolumn{7}{c}{Majority-vote annotation outcome} \\
\midrule
View & Resolved (\%) & Male & Female/Juvenile & Uncertain & Occluded & No majority \\
\midrule
RGB & 32.0 & 463 & 1,873 & 4,102 & 727 & 130 \\
Thermal & 27.3 & 309 & 1,684 & 4,694 & 459 & 149 \\
Matched & 52.8 & 540 & 3,312 & 2,822 & 444 & 177 \\
\bottomrule
\end{tabular}
\end{table}

The annotation data, aligned crops, and the resulting \textit{evaluation pool} with its held-out-flight train, validation, and test splits are available at \url{https://zenodo.org/records/21061638}. Unless otherwise noted, tests use grouped five-fold held-out-flight splits so each flight serves as test once without leakage. The monthly breakdown mirrors the expected antler cycle: RGB struggles around casting and early regrowth, thermal is strongest during growth months but its male labels nearly disappear in winter (3 in January, 0 in March, 0 in December), while matched review stays more stable across those periods, recovering 57 majority-male samples in January, 60 in November, and 13 in December. The single-modality failure periods are therefore not identical, which is precisely why seasonal priors and matched inspection complement each other.

\subsection{Modality testing across uncertainty bands}
\label{subsec:modality_results}

We compare RGB, thermal, and matched features on the \textit{evaluation pool} using a linear SVM, widening and tightening the uncertainty band to trace the accuracy--coverage trade-off; the pool and classifier stay fixed, and only the abstention threshold changes.

\begin{figure}[htbp]
\centering
\includegraphics[width=\linewidth]{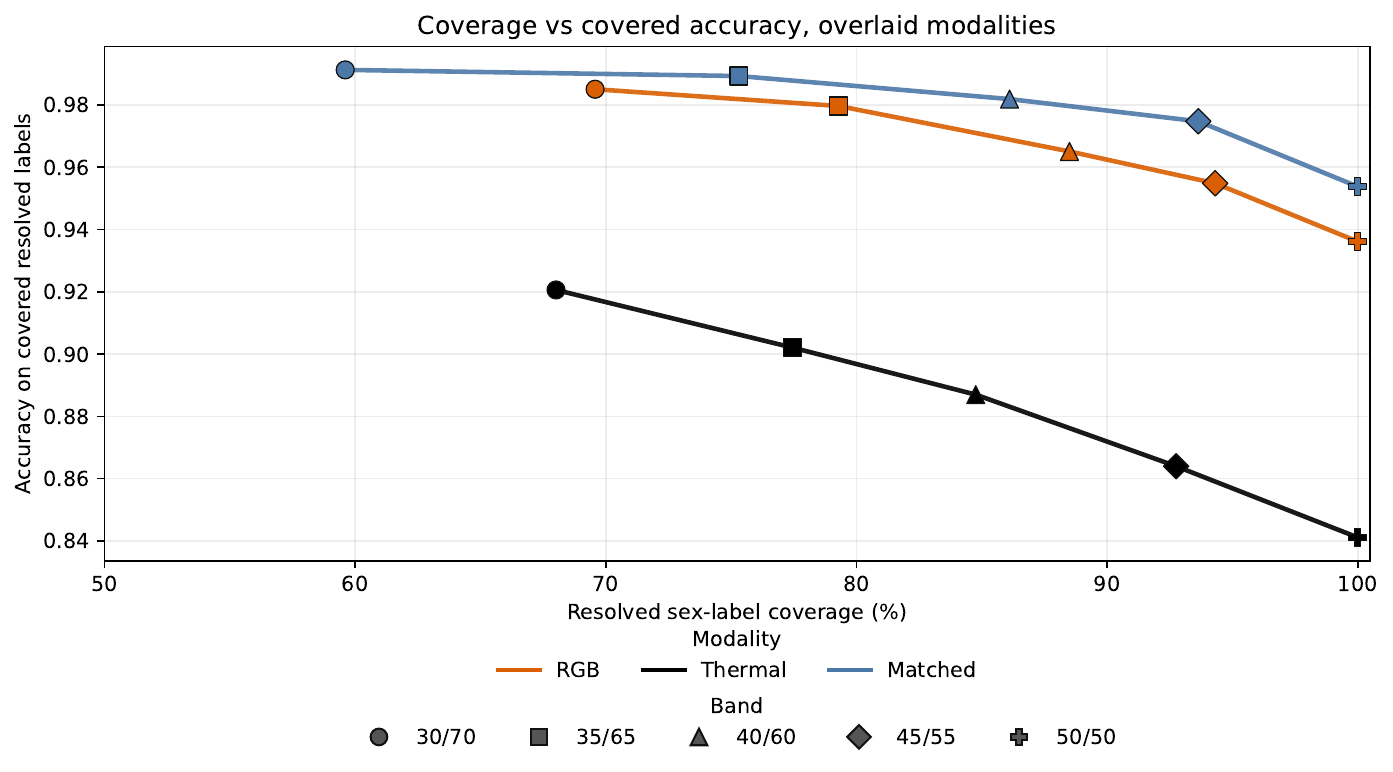}
\caption{SVM accuracy against coverage, where coverage is the fraction of test crops kept outside the uncertainty band. Curves are plotted so that views can be compared at \emph{equal} coverage rather than at a shared band. Fused features give the best accuracy, RGB is consistently second, and thermal remains the weakest view across operating points. Accuracy here is overall accuracy on a female-heavy pool; balanced accuracy and male $F_1$ are reported in \autoref{tab:method_family_comparison}.}
\label{fig:common_pool_modalities}
\end{figure}

\autoref{fig:common_pool_modalities} shows a consistent pattern across operating points. At full coverage every view keeps the same crops, which makes it the equal-coverage comparison: the matched view reaches 95.4\% accuracy and 89.4\% balanced accuracy, against 93.6\%/86.3\% for RGB and 84.1\%/74.0\% for thermal. The fixed 0.35--0.65 band is one operating point, not a like-for-like comparison, since the views land at different coverage: the matched view retains 75.3\% of the test set at 98.9\% covered accuracy, RGB 79.3\% at 98.0\%, thermal 77.4\% at 90.2\%. Fusion buys part of that covered accuracy with four points less coverage than RGB, so banded numbers should be read off the curves at equal coverage. Read at a fixed 80\% coverage the curves give 98.6\% for the matched view, 97.9\% for RGB and 89.7\% for thermal, so holding the three views to the same coverage leaves the ordering unchanged. Because the pool is female-heavy, overall accuracy understates minority-class performance, so we turn to male-focused metrics in \autoref{tab:method_family_comparison}. Of the crops that annotator review left unresolved, 12.5\% fall inside the matched band (10.4\% RGB, 14.9\% thermal), so the model defers on only part of what humans found unresolvable.

\subsection{Classifier-family sensitivity and class-specific abstention}
\label{subsec:classifier_families}

Three classifier families test whether the cross-modal ordering of \autoref{subsec:modality_results} is an artifact of the linear SVM or a property of the representations: a linear SVM, a triplet metric learning head, and a task-trained YOLO image classifier. \autoref{tab:method_family_comparison} gives a qualified yes: matched features are still strongest overall, and among the frozen DINO views RGB remains the strongest single modality while thermal remains the hardest. The qualification matters: RGB $>$ thermal is a property of the \emph{frozen RGB-pretrained DINO representation}, not of the modalities as such. Under the task-trained YOLO26x encoder it reverses, with thermal ahead of RGB by 6.1 accuracy and 25.6 male $F_1$. The SVM and triplet rows share the same frozen features and differ only in the head, and the split protocols differ as set out in \autoref{subsec:models}. Since coverage ranges from 75.3\% to 98.5\% across rows, the full-coverage columns are the like-for-like comparison.

\begin{table}[htbp]
\centering
\caption{Model comparison on the evaluation matched-label pool. Matched features lead under both heads that accept paired input, but the RGB over thermal ordering is a property of the frozen DINO representation only: it reverses under the task-trained YOLO26x encoder, where thermal leads RGB by 6.1 accuracy. `Bal. acc.' denotes balanced accuracy, `M. $F_1$' the male-class $F_1$, and `Cov.' the coverage retained outside the fixed 0.35--0.65 band. `Cov. acc.' and `Cov. M. $F_1$' are computed on that retained subset. YOLO26x is only evaluated for RGB and thermal single-view crops.}
\label{tab:method_family_comparison}
\scriptsize
\setlength{\tabcolsep}{3pt}
\resizebox{\linewidth}{!}{%
\begin{tabular}{@{}l l c c c c c c@{}}
\toprule
View & Method & Acc. & Bal. acc. & M. $F_1$ & Cov. & Cov. acc. & Cov. M. $F_1$ \\
\midrule
RGB & Linear SVM & 93.6 & 86.3 & 77.0 & 79.3 & 98.0 & 90.7 \\
RGB & Triplet metric + centroid & 93.9 & 91.9 & 79.8 & 98.5 & 94.5 & 81.4 \\
RGB & YOLO26x classifier & 86.3 & 65.2 & 41.8 & 94.4 & 87.3 & 43.8 \\
\midrule
Thermal & Linear SVM & 84.1 & 74.0 & 51.4 & 77.4 & 90.2 & 60.9 \\
Thermal & Triplet metric + centroid & 90.3 & 80.3 & 65.1 & 97.1 & 91.0 & 66.3 \\
Thermal & YOLO26x classifier & 92.4 & 77.8 & 67.4 & 96.8 & 93.4 & 70.7 \\
\midrule
Matched & Linear SVM & 95.4 & 89.4 & 83.1 & 75.3 & 98.9 & 94.6 \\
Matched & Triplet metric + centroid & 95.0 & 90.5 & 81.9 & 97.9 & 95.8 & 84.4 \\
\bottomrule
\end{tabular}%
}
\end{table}

The main change across families is in the quality--coverage trade-off rather than the ordering. At the fixed band the triplet head keeps far more samples than the SVM (97.1--98.5\% versus 75.3--79.3\% coverage) but gives up retained accuracy and retained male $F_1$ on RGB and matched views. Thermal is the exception: YOLO26x is strongest there, at 92.4\% accuracy, 67.4\% male $F_1$, and 96.8\% coverage.

\begin{table}[htbp]
\centering
\caption{Class-specific reject rates at the fixed 0.35--0.65 band. Abstention is not class-neutral: the linear SVM defers males more than twice as often as female/juvenile cases (45.9\% versus 21.2\% on matched crops), so its high covered accuracy is partly bought by discarding the minority class, while the triplet and YOLO26x heads narrow the gap. `Gap' is male reject minus female/juvenile reject. YOLO26x is only evaluated for RGB and thermal single-view crops.}
\label{tab:reject_skew}
\scriptsize
\setlength{\tabcolsep}{3pt}
\begin{tabular}{l l c c c c}
\toprule
View & Method & Cov. & Female reject & Male reject & Gap \\
\midrule
RGB & Linear SVM & 79.3\% & 18.2\% & 36.3\% & +18.1\% \\
RGB & Triplet metric + centroid & 98.5\% & 1.2\% & 2.9\% & +1.7\% \\
RGB & YOLO26x classifier & 94.4\% & 6.1\% & 2.0\% & -4.2\% \\
\midrule
Thermal & Linear SVM & 77.4\% & 21.0\% & 32.0\% & +11.0\% \\
Thermal & Triplet metric + centroid & 97.1\% & 2.6\% & 4.9\% & +2.3\% \\
Thermal & YOLO26x classifier & 96.8\% & 3.1\% & 3.9\% & +0.8\% \\
\midrule
Matched & Linear SVM & 75.3\% & 21.2\% & 45.9\% & +24.7\% \\
Matched & Triplet metric + centroid & 97.9\% & 1.7\% & 4.9\% & +3.2\% \\
\bottomrule
\end{tabular}
\end{table}

\autoref{tab:reject_skew} shows why the covered metrics should not be read in isolation. At the fixed band the linear SVM retains the most accurate subset, but it pays for that with the widest male reject gap, and the heads that soften the gap do so mostly by keeping more samples. The headline 98.9\% covered accuracy should not be quoted alone: it is reached at 75.3\% coverage while discarding 45.9\% of male crops. On a female-heavy pool, rejecting males raises covered accuracy almost mechanically, so covered male $F_1$ (94.6\%) and the reject gap belong beside it.

\subsection{Seasonal priors and month-wise uncertainty}
\label{subsec:priors}

We ask whether the seasonal antler calendar aligns with classifier uncertainty, and whether encoding it as sample weights improves the SVM.

\begin{figure}[htbp]
\centering
\includegraphics[width=\linewidth]{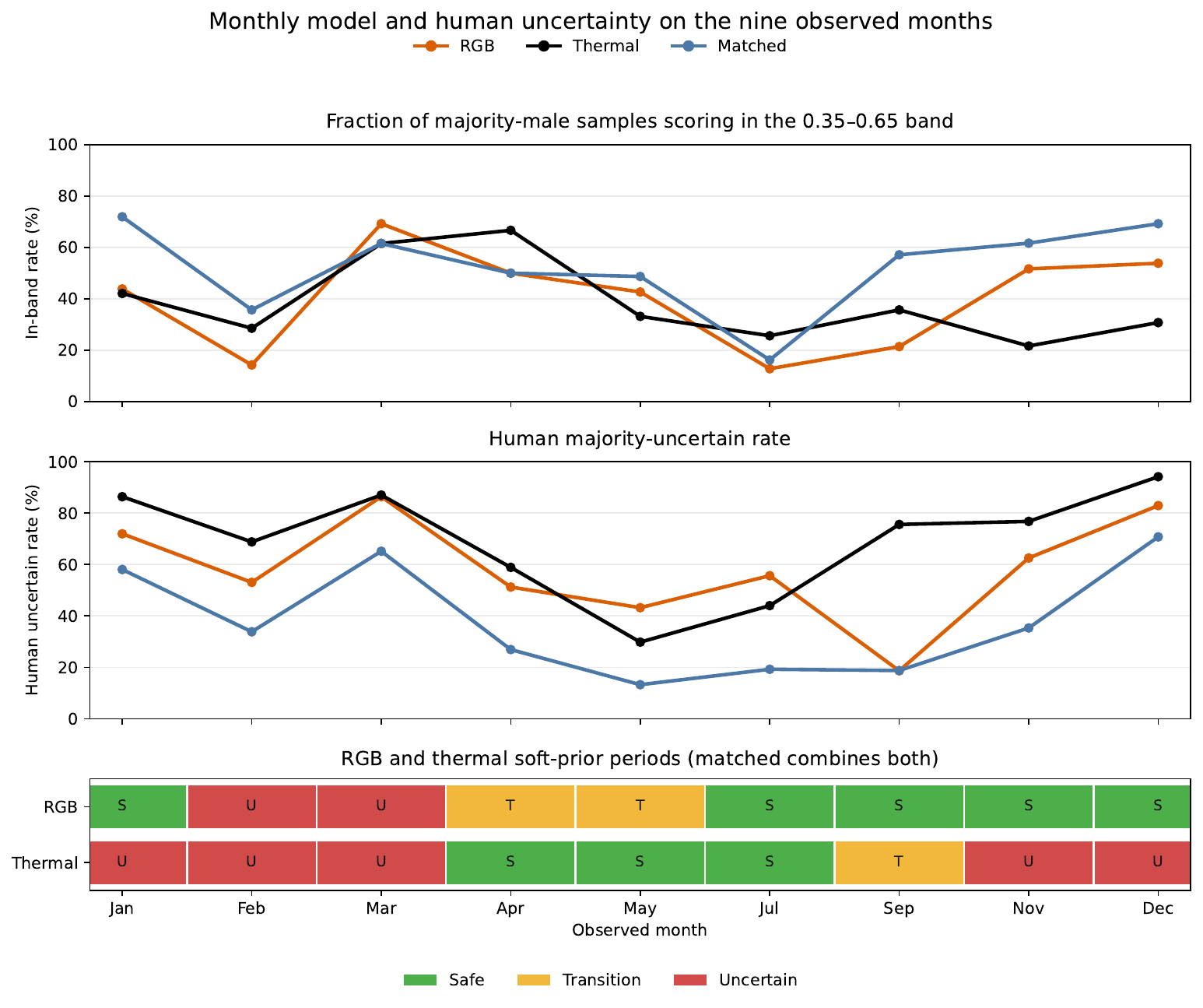}
\caption{Month-wise model and human uncertainty. The top panel shows the fraction of the male deer pool whose score falls inside the fixed 0.35--0.65 band in each month; elevated values mean the model is still unsure on the male cases of interest. The middle panel reports the human majority-uncertain rate by modality and month (both sexes). The bottom strip shows the RGB and thermal soft-prior bins.}
\label{fig:monthly_uncertainty_alignment}
\end{figure}

\paragraph{Month-wise alignment between priors and uncertainty.}
\autoref{fig:monthly_uncertainty_alignment} shows the fraction of male-labeled test crops that fall inside the fixed 0.35--0.65 uncertainty band (top panel), the human majority-uncertain rate by modality and month (middle panel), and the RGB and thermal soft-prior bins (bottom strip). The broad alignment is visible: model uncertainty tracks human annotator uncertainty across months, and both loosely track the prior bins. July, where both modalities are in their ``safe'' prior period, is the least uncertain month averaged across modalities and classifiers. Winter months with uncertain or casting-phase priors show elevated model uncertainty that mirrors the high human abstention rates seen in \autoref{subsec:annotation_results}. The alignment is qualitative and not monotone. Thermal is binned ``safe'' in April yet is most uncertain there, and ``uncertain'' in November and December yet comparatively confident. Month is also confounded with flight, weather and sensor settings, and the interface showed the capture month, so human uncertainty is not blind to the prior.

\paragraph{Effect of prior-aware training weights.} As a quantitative test, we change the per-sample training weight $\alpha_i$ in the linear SVM according to each modality's soft-prior bin, using four schedules over the `safe', `slight', and `uncertain' bins shown in the bottom strip of \autoref{fig:monthly_uncertainty_alignment}: \emph{baseline} $(1.00,1.00,1.00)$, \emph{mild safe boost} $(1.25,1.10,0.90)$, \emph{strong safe boost} $(1.50,1.20,0.60)$, and \emph{safe focus} $(1.75,1.15,0.40)$. Validation and test flights are never reweighted. The effect is small: RGB improves most under `mild safe boost' (+0.2 accuracy, +0.5 male $F_1$), thermal under `strong safe boost' (+0.3, +0.5), and fusion changes least. The gains are consistent in sign but small, so weighting is a secondary use of the calendar rather than a headline result. Comparing season-aware routing against confidence-based routing at equal cost is left to future work.

\paragraph{Channel replacement.} We also ask whether paired sensors require separate embeddings at inference, or can be approximated by injecting thermal intensity into one RGB channel before encoding. \autoref{tab:channel_swap_summary} reports all five variants. Replacing the blue channel is the only one that beats plain RGB, but it still trails late fusion; replacing red or green is worse, which suggests the blue slot is the least disruptive place to inject thermal signal into an RGB-pretrained backbone.

\begin{table}[htbp]
\centering
\caption{Combined modalities and channel-replacement classification scores compared to baseline RGB. Only the blue-channel replacement beats plain RGB (+1.2 accuracy, +4.0 male $F_1$), and it still trails late fusion, so input-level fusion does not remove the need for separate per-modality embeddings.}
\label{tab:channel_swap_summary}
\footnotesize
\setlength{\tabcolsep}{4pt}
\begin{tabular}{@{}l c c c c@{}}
\toprule
View & Acc. & Male $F_1$ & $\Delta$ acc. vs RGB & $\Delta$ male $F_1$ vs RGB \\
\midrule
RGB & 93.6 & 77.0 & 0.0 & 0.0 \\
Thermal $\rightarrow$ R & 93.4 & 75.4 & -0.2 & -1.6 \\
Thermal $\rightarrow$ G & 90.4 & 65.5 & -3.2 & -11.5 \\
Thermal $\rightarrow$ B & 94.8 & 81.0 & +1.2 & +4.0 \\
Matched & 95.4 & 83.1 & +1.8 & +6.1 \\
\bottomrule
\end{tabular}
\end{table}

\section{Discussion and Conclusion}
\label{sec:discussion}

The results point to one underlying structure: the seasonal antler calendar predicts where annotation will be unreliable and where model confidence will be low. Its main use is to guide the annotation protocol and, secondarily, training weights; the abstention rule we evaluate is confidence-based rather than season-aware.

\paragraph{Matched-pair gains and complementary failure modes.}
Matched review roughly doubles the share of crops that can be resolved and returns majority-male labels on hundreds of crops that single-view review left unresolved. Once supervision is fixed, it also yields the strongest downstream representation, consistently across classifier heads. The asymmetry between single views is itself informative: thermal-only review misses far more male labels, consistent with winter and shoulder-season visibility limits, while matched review reverses a smaller set of male calls in the other direction. In thermal, ear position, head tilt, or silhouette artifacts can look antler-like without RGB confirmation; in RGB, poor contrast, blur, or vegetation can hide structure that thermal still shows. Matched review is therefore safer mainly because the two error modes are not shared.

\paragraph{Soft priors and abstention.}
As a diagnostic, the calendar broadly accounts for the months in which annotators and classifiers are most uncertain, subject to the inversions and confounds in \autoref{subsec:priors}. Rejecting predictions inside a fixed band improves covered accuracy but defers a non-random subset of hard cases, so abstention should be reported alongside class-specific reject rates rather than coverage alone.

\paragraph{Designing surveys around identifiability.}
The calendar is also a data collection instrument. Identifiability is concentrated in particular combinations of modality and month, so the cheapest way to improve a classifier is often to fly when the cue is visible rather than to model harder afterwards: hard-antler months for RGB, velvet growth for thermal, and paired capture through casting and early regrowth, when neither view is reliable alone. Survey effort and payload can be planned against that map, and the reasoning extends to any species with seasonal diagnostic cues.

\paragraph{Limitations and future work.}
The binary task still merges females and juveniles, and the priors are hand-specified. The most consequential limitation is that training and evaluation both run on the curated matched majority-vote pool, so gains from curation cannot be separated from gains from the representation. Settling this needs alternative labeling protocols: single-annotator labels, injected label noise, or an independently adjudicated subset. The matched setting is also a late-fusion baseline on frozen DINOv3 features, not an end-to-end model. Useful next steps are stronger fusion, learned priors, thermal-native backbones, and track-level rather than single-crop decisions, where one confident frame may resolve an individual and persistently ambiguous tracks can be flagged for expert review.

{\small
\bibliographystyle{splncs04}
\bibliography{main}
}

\end{document}